\documentclass[sigconf]{acmart}

\usepackage{booktabs} % For formal tables

\setcopyright{rightsretained}
\usepackage{courier}

\usepackage{subcaption}
\usepackage{hyperref}

\acmDOI{10.1145/nnnnnnn.nnnnnnn}

\acmISBN{978-x-xxxx-xxxx-x/YY/MM}

\acmConference[GECCO '19]{the Genetic and Evolutionary Computation Conference 2019}{July 13--17, 2019}{Prague, Czech Republic}
\acmYear{2019}
\copyrightyear{2019}

\acmPrice{15.00}

\begin{document}

\copyrightyear{2019} 
\acmYear{2019} 
\setcopyright{acmlicensed}
\acmConference[GECCO '19]{Genetic and Evolutionary Computation Conference}{July 13--17, 2019}{Prague, Czech Republic}
\acmBooktitle{Genetic and Evolutionary Computation Conference (GECCO '19), July 13--17, 2019, Prague, Czech Republic}
\acmPrice{15.00}
\acmDOI{10.1145/3321707.3321817}
\acmISBN{978-1-4503-6111-8/19/07}
%Alternative: Emerging representations in evolved world models

\title{Deep Neuroevolution of Recurrent and Discrete World Models}

\author{Sebastian Risi and Kenneth O. Stanley}
\affiliation{%
  \institution{Uber AI}
   \city{San Francisco, CA 94103\\
   sebastian.risi@gmail.com,  kstanley@uber.com} 
}
\renewcommand{\shortauthors}{}

\begin{abstract}
Neural architectures inspired by our own human cognitive system, such as the recently introduced world models, have been shown to outperform traditional deep reinforcement learning (RL) methods in a variety of different domains. Instead of the relatively simple architectures employed in most RL experiments, world models rely on multiple different neural components that are responsible for visual information processing, memory, and decision-making. However, so far the components of these models have to be trained separately and through a variety of specialized training methods. This paper demonstrates the surprising finding that models with the same precise parts can be instead efficiently trained \emph{end-to-end}  through a genetic algorithm (GA), reaching a comparable performance to the original world model by solving a challenging car racing task. An analysis of the evolved visual and memory system indicates that they include a similar effective representation to the system trained through gradient descent. Additionally, in contrast to gradient descent methods that struggle with discrete variables, GAs also work directly with such representations, opening up opportunities for classical planning in latent space. This paper adds additional evidence on the effectiveness of deep neuroevolution for tasks that require the intricate orchestration of multiple components in complex heterogeneous architectures.
\end{abstract}

\maketitle

\section{Introduction}
Neuroevolution, i.e.\ evolving neural networks through evolutionary algorithms, has long been applied to complex control problems \cite{stanley2002evolving,floreano2008neuroevolution,risi2017neuroevolution} and has recently been shown to be a competitive alternative for reinforcement learning problems~\cite{such2017deep,salimans2017evolution}. Surprisingly, Such et al. \cite{such2017deep} demonstrated that even a simple genetic algorithm (GA) is able to optimize a large-scale deep network to play various Atari games from raw pixels. 

However, while the aforementioned deep networks have large numbers of parameters, their architectures are often relatively simple feed-forward, directly mapping  high-dimensional inputs to the network's outputs~\cite{such2017deep}. It is therefore an open question how genetic algorithms would scale to problems that require more complex architectures with multiple different and interacting components.

One such neural network-based architecture, which is inspired by the human cognitive system, is the world model recently introduced by Ha and Schmidhuber~\cite{ha2018recurrent}. This agent model contains three different components: a visual module that maps high-dimensional inputs to a lower-dimensional representative code, a memory component that tries to predict the future based on past experience, and a decision-making module that determines the action of the agent based on inputs from the visual and memory module.   

The world model is motivated by the insight that our brains learn abstract representations of both spatial and temporal data, allowing us to generalize to different situations and to predict potential future sensory experiences. Because of its predictive abilities, the world model approach is able to find a solution for a challenging 2-D car racing task (defined as reaching a minimum average reward of 900 over 100 consecutive trials), a domain that other deep RL methods such as Q-Learning and A3C \cite{khan2018,jang2017} struggle with so far. However, the approach requires each of its three components to be trained separately and through specialized training methods. While the controller part is trained through an evolution strategy, both the visual and memory components are trained through stochastic gradient descent based on random rollouts. Given the surprising and competitive results of GAs on RL problems \cite{such2017deep}, the question in this paper is whether a simple GA might also be competitive with complex heterogeneous systems like world models, and if so, what type of representation would evolve.  

As the results in this paper on a 2-D car racing domain demonstrate, it is in fact possible to train a complex multi-component system end-to-end with a simple genetic algorithm. Indeed, the GA performs comparably to the world model approach and finds a solution to the task, outperforming all of the other traditional deep RL methods. Surprisingly, even though the sensory component was not directly trained to compress similar sensory states to similar latent codes (as is the case in the training of the autoencoder in Ha and Schmidhuber~\cite{ha2018recurrent}), the GA discovers  such a representation by itself because it is beneficial for solving the task. Similarly, the emergent representation of the memory system is able to  predict situations in which the agent needs to react quickly to changes in the environment, such as when taking sharp turns.

Additionally, this paper introduces a discrete world model approach, in which the VAE is restricted to binary outputs. While traditional machine learning techniques have focused on continuous representations  because backpropagating through discrete variables is challenging \cite{bengio2013estimating,raiko2014techniques,rolfe2016discrete,van2017neural}, evolutionary-based approaches do not struggle with discrete representations. In the future, such representations could directly support classical planning approaches in latent space \cite{asai2018classical}. 

Overall, the performance of the GA for evolving the weights of more complex  architectures suggests that it can be a competitive alternative in many tasks that were thought too high-dimensional for artificial evolution. In the future, it will be interesting to extend this approach to not only evolving the network's weights but also the architectures of the world models themselves. 

\section{Related Work}
A variety of different RL algorithms have recently been shown to work well on a diverse set of problems when combined with the representative power of deep neural networks \cite{mnih2015human,schulman2015trust,schulman2017proximal}. While most approaches are based on variations of Q-learning \cite{mnih2015human} or policy gradient methods \cite{schulman2015trust,schulman2017proximal}, recently evolutionary-based methods have emerged as a competitive alternative  \cite{such2017deep,salimans2017evolution}. 
\begin{figure*}[ht]
\centering
\includegraphics[width=1.0\textwidth]{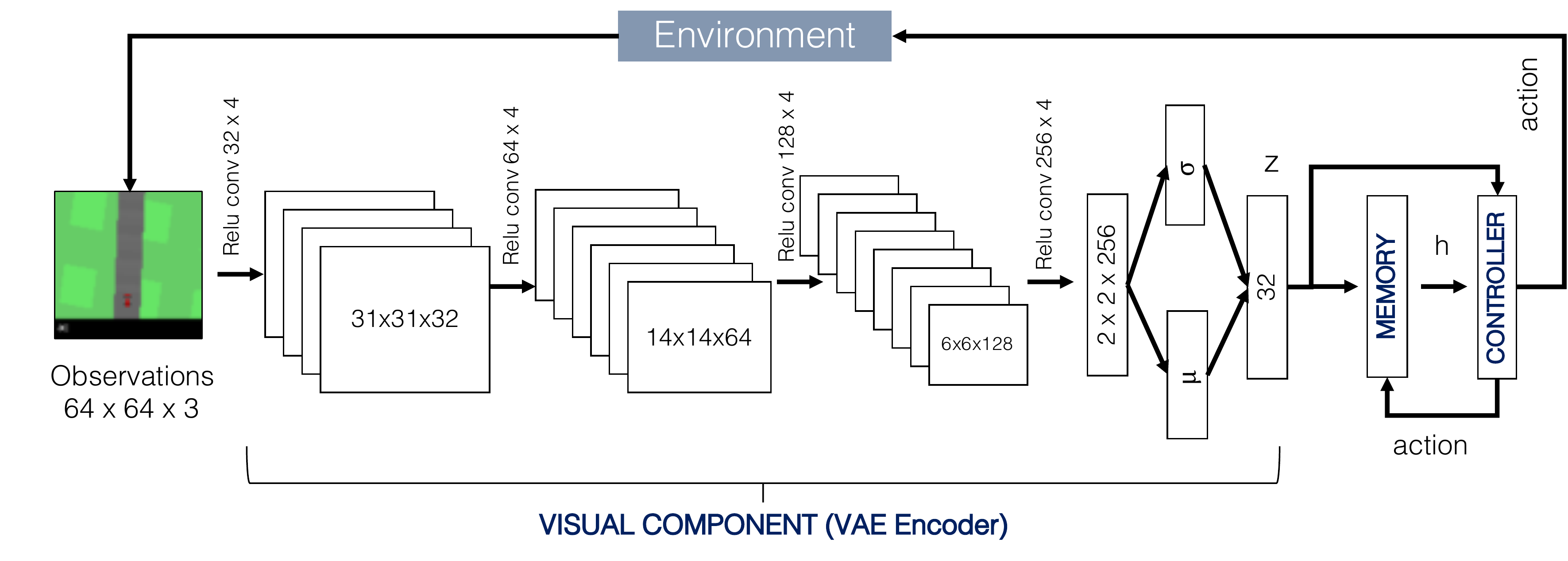} \vspace{-0.2in}
\caption{Agent Model. \normalfont The agent model consists of three modules. A visual component (the encoder of a variational autoencoder) produces a latent code $z_t$ at each time step $t$, which is concatenated with the hidden state $h_t$ of the LSTM-based memory component that takes $z_t$ and previous  performed action $a_{t-1}$ as input. The combined vector ($z_t, h_t$) is input into the controller component to determine the next action of the agent. In this paper, the agent model is trained end-to-end with a genetic algorithm.} 
\label{fig:network}
\end{figure*}

Salimans et al.~\cite{salimans2017evolution} showed that a type of evolution strategy (ES) can reach competitive performance in the Atari benchmark and at controlling robots in MuJoCo. Additionally, Such et al.~\cite{such2017deep} demonstrated that a simple genetic algorithm is in fact able to reach similar performance to  deep RL methods such as DQN or A3C.  Earlier approaches that evolved neural networks for RL tasks worked well in complex RL tasks with lower-dimensional input spaces \cite{stanley2002evolving,floreano2008neuroevolution,risi2017neuroevolution} and also showed promise in directly learning from high-dimensional input \cite{koutnik2014evolving}.

However, when trained end-to-end  these networks are often still orders of magnitude simpler than networks employed for supervised learning problems \cite{justesen2019deep} or depend on additional losses that are responsible for training certain parts of the network \cite{wayne2018unsupervised}. 

More complex agent models often require training different network components separately \cite{wahlstrom2015pixels, ha2018recurrent}. For example, in the world model approach \cite{ha2018recurrent}, the authors first train a variational autoencoder (VAE) on 10,000 rollouts from a random policy to compress the high-dimensional sensory data and then train a recurrent network to predict the next latent code. Only after this process is a smaller controller network trained to perform the actual task, taking information from both the VAE and recurrent network as input to determine the action the agent should perform.

In another earlier related approach the authors first train an autoencoder in an unsupervised way \cite{alvernaz2017autoencoder} or train an object recognizer in a supervised way \cite{poulsen2017dlne} and then in a separate step evolve a controller module. The idea in the present paper is to explore whether a GA can optimize a multi-component system end-to-end without the need to separate training into different phases, which is explained in the next section.

Approaches to learning dynamical models have mainly focused on gradient descent-based methods, with early work on RNNs in the 1990s \cite{schmidhuber1990line}. More recent work includes PILCO \cite{deisenroth2011pilco}, which is a probabilistic model-based policy search method and Black-DROPS \cite{chatzilygeroudis2017black}, which employs CMA-ES for data-efficient optimization of complex control problems. Additionally, interest has increased in learning  dynamical models directly from high-dimensional pixel images for robotic tasks \cite{watter2015embed}  and also video games \cite{guzdial2017game}. Work on evolving forward models has mainly focused on neural networks that contain orders of magnitude fewer connections and lower-dimensional feature vectors   \cite{norouzzadeh2016neuromodulation} than the models in this paper. 

\section{End-to-end Training of Multi-Component Networks}
The agent model in this paper is based on the world model approach introduced by Ha and Schmidhuber~\cite{ha2018recurrent}. The network  includes a sensory component, implemented as a variational autoencoder (VAE) that compresses the high-dimensional sensory information into a smaller 32-dimensional representative code (Figure~\ref{fig:network}). This code is fed into a memory component based on a recurrent LSTM~\cite{hochreiter1997long}, which should predict future representative codes based on previous information. Both the output from the sensory component and the memory component are then fed into a controller that decides on the action the agent should take at each time step. 

Following Such et al.~\cite{such2017deep}, the  deep neural networks are evolved with a simple genetic algorithm, in which mutations add Gaussian noise to the parameter
vectors of the networks. Three different mutation operators are investigated:
\begin{itemize}
\item In the first approach (\textbf{MUT-ALL}), we apply additive Gaussian noise to the parameter vectors of all three modules (vision, memory, and controller) at the same time: $ \theta' = \theta + \sigma \epsilon $, where $	\epsilon \sim N (0, I)$ and $\sigma$ was determined empirically and set to $0.01$ for the experiments in this paper. 
\item In the second module mutation setup (\textbf{MUT-MOD}), a mutation has an equal probability to either mutate the visual, memory, or controller component of the network. The hypothesis is that this treatment should allow evolution to better fine-tune each component in the system than an approach that always adds Gaussian noise to all components. 
\item To elucidate the advantages of evolving both the VAE and memory component, their weights are randomly chosen in the third setup (\textbf{MUT-C}) and only the controller component is modified through evolution.    
\end{itemize}

In the original world model approach the visual and memory component were trained separately and through unsupervised learning based on data from random rollouts. Here they are optimized through a simple genetic algorithm and the components are not evaluated individually. In other words, the VAE is not directly optimized to reconstruct the original input data and neither is the memory component  optimized to predict the next time step; the whole network is trained in an end-to-end fashion and has to learn a representation by itself that allows it to solve the given task. 

Another potential benefit of GAs, beyond being able to train the whole system end-to-end, is that training discrete VAEs, in which the latent code takes on only binary values, are seamlessly supported. While learning representations  with continuous features have been the focus in machine learning, discrete VAEs can have benefits for domains that are composed of discrete elements (such as language) or can naturally support classical planning approach in latent space \cite{asai2018classical}. However, discrete VAEs have proven difficult to train through gradient descent-based methods \cite{bengio2013estimating,raiko2014techniques} or require a more complicated training procedure \cite{rolfe2016discrete,van2017neural}, because backpropagating through discrete variables is not directly possible. This paper tests the idea of evolving discrete VAEs for the car racing domain. The  \textbf{DISCRETE-MOD} approach feeds the original output of the VAE encoder  through a step function that maps the continuous outputs to binary values. 
\section{Experiment}

\begin{figure}
\centering
	\begin{subfigure}[b]{0.24\textwidth} 
\centering
\includegraphics[width=0.9\textwidth]{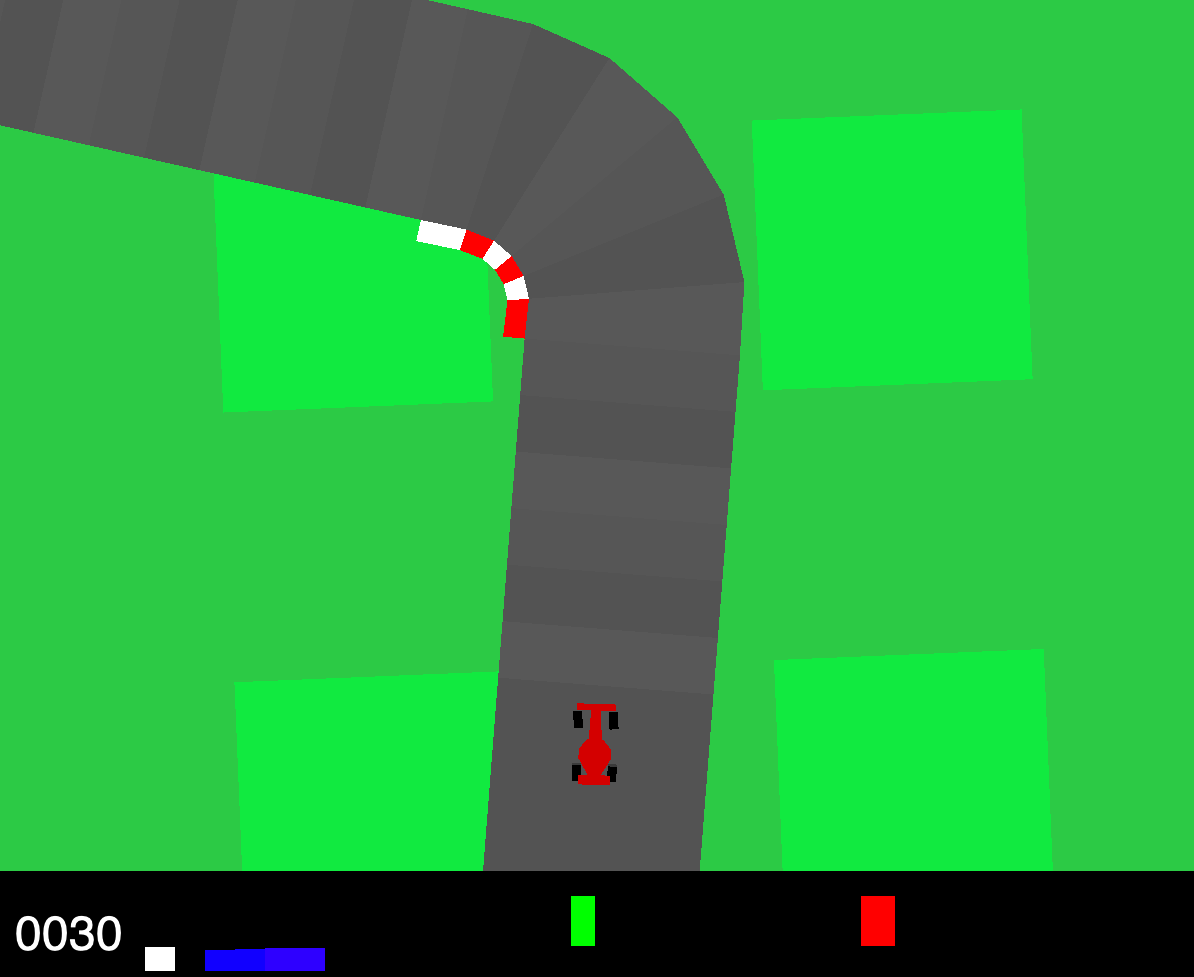}
\caption{Original}
\end{subfigure} 
	\begin{subfigure}[b]{0.195\textwidth} 
\centering
\includegraphics[width=0.9\textwidth]{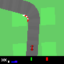}
\caption{Scaled \normalfont ( $64 \times 64 \times 3)$} 
\end{subfigure} 
\caption{Car Racing Domain. \normalfont In the car racing domain the agent has to learn to drive across many procedurally generated tracks as fast as possible from $64 \times 64$ RGB color images.}
\label{fig:car_domain}
\end{figure}

Following the original world model approach~\cite{ha2018recurrent}, in the experiments in this paper an agent is trained to solve a challenging 2-D car racing tasks from 64$\times$64 RGB pixel inputs (Figure~\ref{fig:car_domain}). In this continuous control  task \texttt{CarRacing-v0}~\cite{oleg2016} the agent is presented with a new procedurally generated track every episode,  receiving a reward of -0.1 every frame and a reward of +100/$N$ for each visited track tile, where $N$ is the total number of tiles in the track.  The network controlling the agent (Figure~\ref{fig:network}) has three outputs to control left/right steering, acceleration and braking. Further details on the network model can be found in Section~\ref{sec:setup}. 

Training agents in procedurally generated environments  \cite{togelius2011search}, instead of only a particular one, can significantly increase their generality in a variety of different domains and prevent overfitting  \cite{justesen2018illuminating,cobbe2018quantifying,zhang2018study}. 
Because each agent is tested on a new randomly created track each evaluation, we evaluate the top three individuals of each generation 20 times and average the results to get a better estimate of the true elite (which gets assigned a fitness of $\infty$). Individuals for the next generation are composed of the 50\% highest performing individuals in the current generation plus their offspring determined stochastically through 2-way tournament selection. No crossover operation was employed. To reduce the computational resources spent on non-promising individuals, an evaluation is terminated early in the experiments reported here if an agent is not able to reach an unvisited track tile in 20 time steps. 

Following Ha and Schmidhuber~\cite{ha2018recurrent}, after training the champions found in each generation are evaluated on 100 randomly created tracks to estimate their generalization abilities. While it is not difficult to learn to drive slowly around a track, it is challenging to find a solution that can drive  around any given track perfectly and as fast as possible. In fact, many  traditional deep RL methods \cite{khan2018,jang2017}, which additionally also require  pre-processing such as edge detection \cite{jang2017} or stacking recent frames \cite{khan2018,jang2017}, fail to reach high scores on this task (also see Table~\ref{table:scores}). Interestingly, Ha and Schmidhuber showed that a world model without the recurrent memory model receives a significantly lower score in this domain (decreasing from an average of 906$\pm$21 to 788$\pm$141), displaying more unstable driving behaviors. This result suggests the importance of a  memory model in predicting potential futures that allow the agent to take sharp corners seamlessly. An interesting question is whether evolution will discover such dynamics by itself without being explicitly rewarded to doing so. 

\begin{table}
\centering
\caption{Number of parameters and training procedures. \normalfont The visual component of the agent (see Figure~\ref{fig:network}) is effectively only utilizing  and evolving the encoder part of the VAE, which has 755,744 parameters. The decoder network is composed of four deconvolutional layers and has 3,592,803 parameters.}
\begin{tabular}{|l|l|l|l|} 
\hline
\textbf{Model}               & \textbf{\#Params}  & \textbf{WM Training} \cite{ha2018recurrent} & \textbf{GA Training}             \\ 
\hline
VAE                  & 4,348,547 & SGD - 1 epoch        &                  \\ 
\cline{1-3}
MD-RNN               & 384,071   & SGD - 20 epochs      & Pop size 200     \\ 
\cline{1-3}
Controller           & 867       & CMA-ES - Pop 64~     & Rollouts 1~ ~      \\ 
\cline{1-2}
\multicolumn{1}{l}{} &           & Rollouts 16            & Solved: 1,200  \\
\multicolumn{2}{l|}{}            & Solved: 1,800 Gen.      &                  \\
\cline{3-4}
\end{tabular}
\label{lbl:table} 
\end{table}

\subsection{Experimental Setup and Model Details}
\label{sec:setup}
%https://pytorch.org/docs/stable/_modules/torch/nn/init.html
%
The size of each population is 200 and evolutionary runs have a termination criterion of 1,000 generations. An overview of the agent model is shown in Figure~\ref{fig:network}, which employs the same architecture as the original world model approach \cite{ha2018recurrent}. The sensory model is implemented as a variational autoencoder that compresses the high-dimensional input to a latent vector $z$. The VAE takes as input an RGB image of size $64 \times 64 \times 3$, which is passed through four convolutional layers, all with stride 2. Details on the encoder are depicted in the visual component shown in Figure~\ref{fig:network}, where layer details are shown as:  activation type (e.g.\ ReLU), number of output channels $\times$ filter size. The decoder, which is in effect only used to analyze the evolved visual representation in Section~\ref{sec:encoder}, takes as input a tensor of size $1 \times 1 \times104$ and processes it through  four deconvolutional layers each with stride 2 and sizes of $128 \times 5$, $64 \times 5$, $32 \times 6$,  and $32\times 6$. The network's weights  are set using the default PyTorch initilisation (He initialisation \cite{he2015delving}), with the resulting tensor being  sampled from $\mathcal{U}(-\text{bound}, \text{bound})$, where $\text{bound} = \sqrt{\frac{1}{\text{fan\_in}}}$.

The memory model \cite{ha2018recurrent} combines a recurrent LSTM network with a mixture density Gaussian model as network outputs, known as a MDN-RNN~\cite{ha2017neural,graves2013generating}. The network has 256 hidden nodes and models $P(z_{t+1} | a_{t}, z_{t}, h_{t})$, where $a_{t}$ is the action taken by the agent at time $t$ and $h_t$ is the hidden state of the recurrent network. Similar models have previously been used for generating sequences of sketches \cite{ha2017neural} and handwriting \cite{graves2013b}. The controller component is a simple linear model that directly maps   $z_t$ and $h_t$ to actions: $a_t = W_c [z_t h_t] + b_c,$ where $W_c$ and $b_c$ are weight matrix and bias vector. Table~\ref{lbl:table} summarizes the parameter counts of the different world model components and how they are trained here and in the world model paper.  The code for the experiments in this paper can be found at: \texttt{\url{https://github.com/sebastianrisi/ga-world-models}}. It is build upon a PyTorch reimplementation of the  world model paper by Tallec et al.~\cite{tallec2018pytorch}.

\section{Results}
\begin{figure}[h]
\centering
\includegraphics[width=0.5\textwidth]{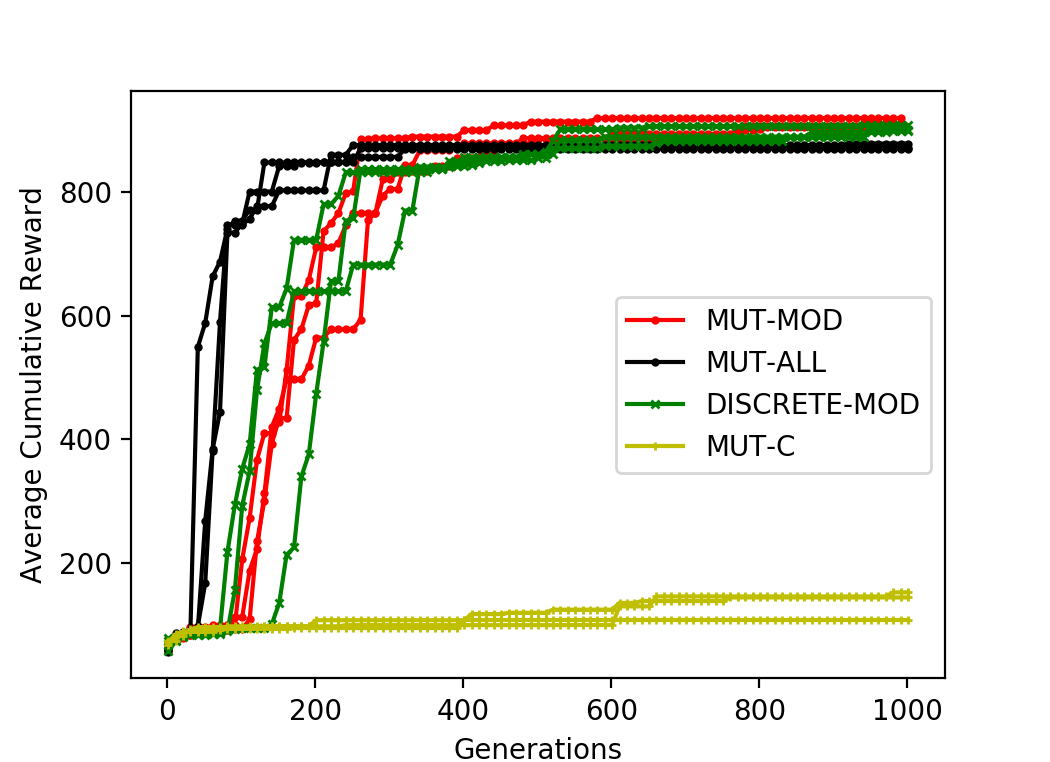}
\caption{Training performance on \texttt{CarRacing-v0}. \normalfont  The score of the best individual is shown in each generation evaluated on 20 randomly created tracks. All approaches, except MUT-C, are able to evolve agent models that can drive around the track at high speed while making very few mistakes.}
\label{fig:fitness}
\end{figure}

Figure~\ref{fig:fitness} shows the performance of each treatment for three independent evolutionary runs. Each evolutionary run took approximately two days to train on a 32-core CPU machine. Mutating either every parameter in the network or only targeting specific modules does not result in large changes although there are some notable differences. While MUT-ALL initially increases faster than MUT-MOD, all three runs of the latter ultimately reach a higher performance than any of the MUT-ALL runs. This result suggests that  it can initially be beneficial to change many parameters at the same time to get a  rudimentary behaviour but fine-tuning them is easier with a mutation operator that only changes one module at a time. The discrete VAE version DISCRETE-MOD also reaches a similar performance to the other methods, confirming the hypothesis that a GA can seamlessly learn a discrete representation that is useful for the task at hand. The results also demonstrate that only mutating the controller part of the neural architecture (MUT-C) and relying on the  features produced by a randomly initialized VAE and MDN-RNN are not enough to allow the agent to learn to drive. 

Interestingly, separate evolutionary runs often follow similar performance curves (which we also observed in  other experiments). This behaviour appears very different from the training of networks with orders of magnitude fewer parameters traditionally studied in neuroevolution, which often have a higher variance across runs  \cite{stanley2002evolving,floreano2008neuroevolution,risi2017neuroevolution}. Analyzing this phenomenon in more detail is an interesting future research direction that we aim to investigate. 
% 
%Average 903.4378011956785 72.13375734037172. %Genome: best_1_0_G1269.p

After 1,000 generations the MUT-MOD agents were getting very close to solving the domain, learning to drive around the track very effectively with few errors, with the best network reaching a generalization score of 888 $\pm$ 66. Therefore we continued evolution for another 200 generations with a lower mutation rate of 0.003 and evaluated each elite on 40 instead of 20 trials. This approach led to finding a solution to the task that reached a score of 903 $\pm$ 72. This average score is comparable to the original world model paper and higher than any traditional RL approaches, which reach scores of around 591 to 893 on average (Table~\ref{table:scores}). A video of the best agent driving around the track can be found at: \url{https://youtu.be/a-tcsnZe-yE}.

\begin{table}[]
\centering
\caption{\normalfont \texttt{CarRacing-v0} scores of different approaches. Only the original world model and the GA approach introduced in this paper are able to solve the task (reaching an average score over 900). }
\begin{tabular}{ll}
\hline
Method      & Average Score \\ \hline
DQN \cite{prieur2017}       &     343 $\pm$ 18   \\
DQN + Dropout \cite{gerber2018}  &     893 $\pm$ 41   \\
A3C (Continious) \cite{jang2017}  &     591$ \pm$ 45   \\
A3C (Discrete)  \cite{khan2018}  &  652 $\pm$ 10  \\
CEOBILLIONAIRE (Gym leaderboard) & 838 $\pm$ 11            \\ 
World model \cite{ha2018recurrent} &  \textbf{906}  $\pm$ 21 \\ 
World model with random MDN-RNN \cite{tallac2018} &  870  $\pm$ 120 \\ 
GA (ours) & \textbf{903} $\pm$ 72 \\ \hline
\end{tabular}
\label{table:scores}
\end{table}

\subsection{Learned Visual Encoder Representation}
\label{sec:encoder}
Because the visual component in our experiments is not specifically trained to reconstruct the given sensory input, it is interesting to analyze what information is contained in the learned latent vector representation. To analyze this question, the evolved VAE encoder weights of a champion network are kept fixed while a decoder is trained in a unsupervised way  to reconstruct data collected from a random policy. The decoder is trained for 100 iterations with the Adam optimization algorithm, a learning rate of 0.0001, using the mean squared distance between the reconstructed image and the input image as loss, in addition to Kullback-Leibler (KL) loss.

\begin{figure}
\centering
	\begin{subfigure}[b]{0.22\textwidth} 
\includegraphics[width=0.9\textwidth]{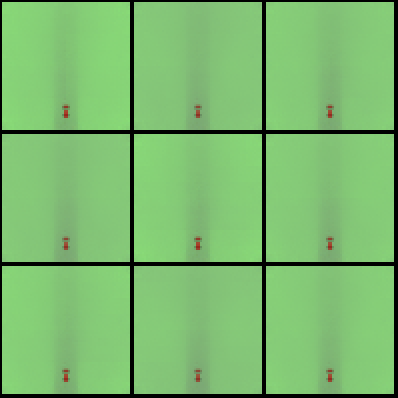}
\caption{Random Weights}
\end{subfigure} \begin{subfigure}[b]{0.22\textwidth}
\includegraphics[width=0.9\textwidth]{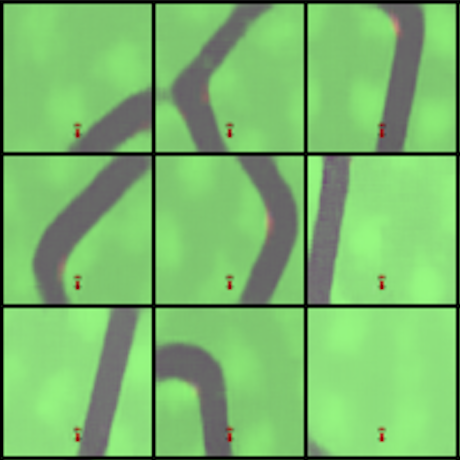}
\caption{Evolved Weights }
\end{subfigure}
\caption{VAE Reconstructions. \normalfont While a network with random encoder weights does not allow the VAE to learn to reconstruct the given images (a), the evolved weights of a champion network encoder enabled different road images to be reconstructed (b). These results suggest that evolution discovered how to compress useful information in the latent code produced by the VAE's encoder.}
\label{fig:vae_samples}
\end{figure}

Interestingly, while the initial random networks from the first generation do not allow the reconstruction of different track images (Figure~\ref{fig:vae_samples}a), which suggests that the initial random weights fail to capture some important information from the pixel inputs, the latent code of the evolved representation contains enough information for this task  (Figure~\ref{fig:vae_samples}b). However, the low reconstruction error does not directly explain how the evolved network is utilizing this compressed representation to drive efficiently around the track. 
\begin{figure*}
\centering
\includegraphics[width=0.75\textwidth]{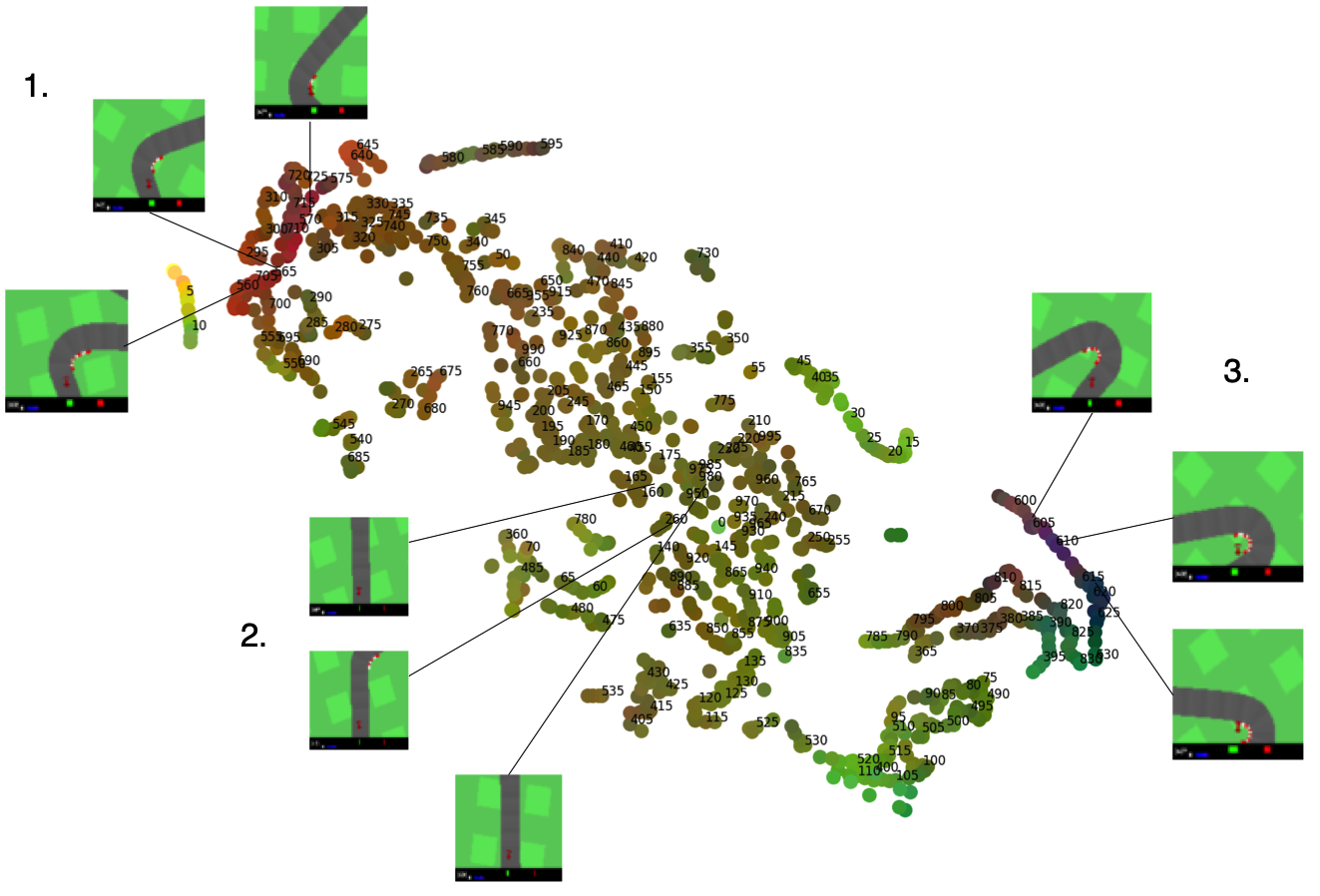}
\caption{Evolved Visual Representation. \normalfont t-SNE mapping of the 32-dimensional latent vectors onto two dimensions. The three action outputs of the agent are mapped to the RGB color values of each plot point (R=steer, G=gas, B=break). The GA successfully discovered a visual encoder that maps similar pixel inputs to similar latent codes. Similar latent codes in turn determine similar agent actions, such as turning right (1), driving straight (2), or turning left (3).}
\label{fig:tsne}
\end{figure*}

To analyze this question further, we employ the t-SNE dimensionality reduction technique \cite{maaten2008visualizing} to map the sequence of 32-dimensional latent vectors collected during one car racing rollout to two dimensions. t-SNE has already been proven valuable for gaining insight into the workings of deep neural networks \cite{such2018atari,mnih2015human}.  

\begin{figure*}
\centering
%\begin{subfigure}[b]{0.25\textwidth} 
\includegraphics[width=0.6\textwidth]{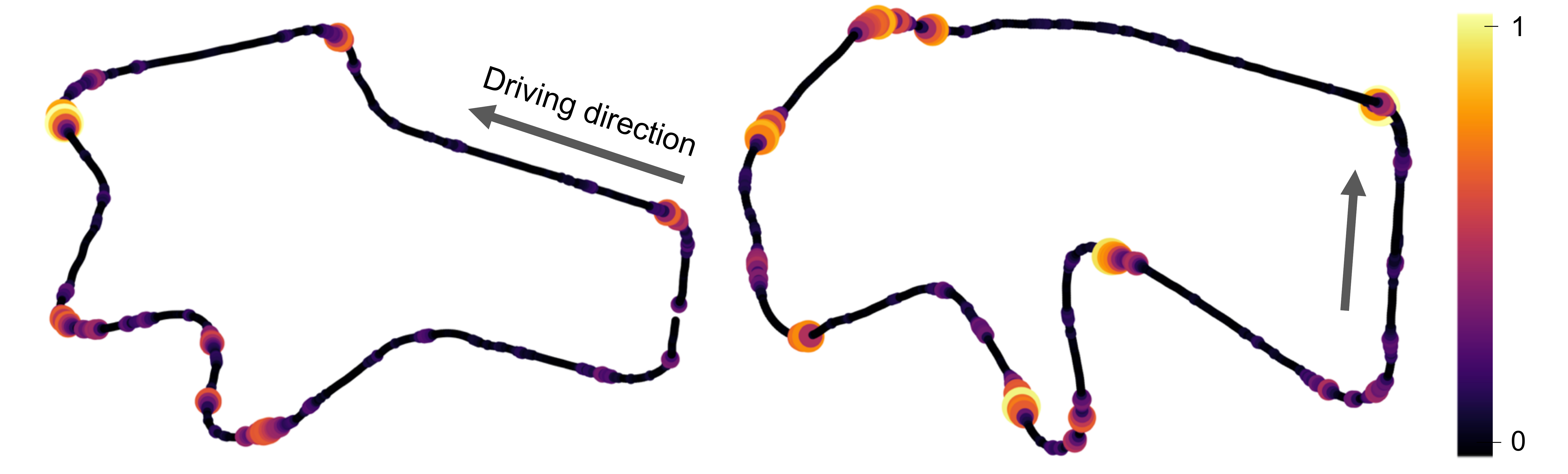}
%\end{subfigure}\begin{subfigure}[b]{0.25\textwidth}
%\includegraphics[width=1.0\textwidth]{images/track1_lstm.png}
%\end{subfigure}
\caption{Dynamics of Evolved Forward Model. \normalfont The size and color of each marker reflects the variance in activation levels of the MDN-RNN's hidden state while the agent is driving around the track (brighter colors and larger marker sizes indicate higher magnitudes). Variance  levels are typically higher when the agent is near corners, while they are lower on straight road segments. These results suggest that the evolved agent model is most reliant on the memory component in situations that benefit from predicting future sensory states.}
\label{fig:forward}
\end{figure*}

The mapping (Figure~\ref{fig:tsne}) suggests that the agent learned to represent situations that require a similar action (e.g.\ turning sharp left or right) with a similar latent vector. For example, situations in which the agent needs to turn right are represented by similar latent codes, which are clustered together, while latent codes for situations in which the agent needs to drive straight or turn left are part of a different cluster. By learning an abstract, compressed representation of the higher-dimensional pixel inputs, it becomes easier for the controller module to learn the required behaviors. 

\subsection{Learned Forward Model Dynamics}
In addition to the visual encoder representation, it is interesting to investigate the emergent dynamics of the evolved predictive memory component.

Figure~\ref{fig:forward} visualizes the activation levels of the MDN-RNN while the agent is driving around two tracks. To get a better sense of the dynamics of the system, we are interested in how much the average activation $x_t$ of all 256 hidden nodes at time step $t$ differs from the overall average across all time steps $\bar{X}=  \frac{1}{N} \sum_1^N \bar{x_t}$. The variance of $\bar{x_t}$ is thus calculated as $\sigma_t = (\bar{X}-\bar{x_t})^2 $, and normalized to the range $[0, 1]$ before plotting. Activation levels far from the mean should have a higher impact on the agent's controller component and can indicate situations in which the agent needs to pay particular attention to the predictions of the MDN-RNN.

The results show that the dynamics of the recurrent network are changing more drastically when the agent is near a corner and change less when the agent is driving on straight track segments. This effect confirms  the hypothesis that predicting future sensory states is particularly important during situations in which the agent needs to react quickly to changes in the environment. 

\section{Discussion and Future Work}
This paper demonstrated that genetic algorithms can not only train the weights of relatively simple network architectures but also complicated systems with over a million weights that include different components for sensory processing, memory, and decision making in an end-to-end fashion. The approach outperforms standard deep RL approaches and reaches a comparable performance to the recently-introduced world model approach that relies on a much more complex training regimen. Another surprising result is that the GA found a solution with a  population size of only 200, compared to the much larger population sizes of 1,000 in the work by Such et al.~\cite{such2017deep} on Atari video game playing. 

The difference between the two mutation treatments MUT-MOD and MUT-ALL also suggests a potentially useful hybrid approach, in which mutations sometimes affect all network layers and sometimes only one layer at a time. Such an approach could combine the better initial exploration of MUT-ALL with the ability of MUT-MOD to better fine-tune different parts of the network in later generations.

While the final average generalization score is slightly lower than the score in the original world model paper \cite{ha2018recurrent}, the presented system does indeed solve the task while not relying on first learning from a large number of random rollouts; instead the system can learn  directly in interaction with the environment. The slightly lower average score (903 compared to 906) with a higher standard deviation (72 compared to 21) could be explained by the fact that if random rollout data is available, training each component separately might produce slightly more robust solutions. However, especially in more complicated tasks for which data collected during random rollouts is insufficient (because a random rollout might not reach all relevant parts of the environment), the end-to-end learning approach could become more important. 

One advantage and indeed motivation of the original world model approach was the fact
that agents can  train and improve in the environments generated by the world model itself, without using the actual  environment. Testing the evolved world model presented in this paper for the same purpose is an important next step. As noted by Ha and Schmidhuber \cite{ha2018recurrent}, the discrete modes in the mixture density model can be beneficial in  environments with random discrete events (e.g.\ firing of a weapon in an FPS game). They observed that if the temperature parameter that controls the model's uncertainty is set to a very low value, the enemies in the world model of their FPS environment never fire their weapons; the MDN-RNN is not able to reach a mode in the mixture of Gaussian models in which this event happens. In this context, we hypothesise that the ability of the GA to evolve discrete VAE representations (which are fed into the MDN-RNN) could make it even easier for the model to switch between different modes than the current continuous VAE  version. 
 
Another exciting prospect is not only to evolve the weights of such large-scale deep networks but also the neural architectures themselves. While evolutionary algorithms have allowed the architectures of relatively simple networks to be evolved for reinforcement learning problems, so far larger-scale architectures have mostly  been evolved in combination with supervised learning \cite{miikkulainen2019evolving} and not extended to very  complex RL problems. Other promising extensions to the simple GA used in this paper could be  additional crossover operators, indirect encodings such as HyperNEAT \cite{stanley2009hypercube,risi2012enhanced}, safe mutations \cite{lehman2017safe}, or more exploratory search methods such as novelty search \cite{lehman2008exploiting}. 

\vspace{-0.05in}
\section*{Acknowledgements}
We would like to thank the anonymous reviewers and David Ha for their insightful comments that very much improved the presentation of this paper. We would also like to thank all of the members of Uber AI for helpful discussions. 
\vspace{-0.05in}
\bibliographystyle{ACM-Reference-Format}

\end{document}